\documentclass{article}
\usepackage{graphicx}
\usepackage{amsmath,stackengine}
\usepackage{multirow}
\usepackage{makecell}
\renewcommand\lbrack[1][]{\ooalign{\ensuremath{#1\lfloor}\cr\ensuremath{#1\lceil}}}
\renewcommand\rbrack[1][]{\ooalign{\ensuremath{#1\rfloor}\cr\ensuremath{#1\rceil}}}

\usepackage{tikz}
\def\checkmark{\tikz\fill[scale=0.4](0,.35) -- (.25,0) -- (1,.7) -- (.25,.15) -- cycle;} 

\usepackage[preprint]{amlc-rq-2022}

\usepackage[utf8]{inputenc} 
\usepackage[T1]{fontenc}    
\usepackage{hyperref}       
\usepackage{url}            
\usepackage{booktabs}       
\usepackage{amsfonts}       
\usepackage{nicefrac}       
\usepackage{microtype}      

\title{MRQ :  Support Multiple Quantization Schemes through Model Re-Quantization}

\author{%
  Manasa Manohara \\
  Amazon Lab126 \\
  Sunnyvale, CA 94089 \\
  \texttt{manomana@amazon.com} \\
   \And
  Sankalp Dayal \\
  Amazon Lab126 \\
  Sunnyvale, CA 94089 \\
  \texttt{sankalpd@amazon.com} \\
  \And
  Tariq Afzal\\
  Amazon Lab126 \\
  Sunnyvale, CA 94089 \\
  \texttt{tafzal@amazon.com} \\
  \AND
  Rahul Bakshi\\
  Amazon Lab126 \\
  Sunnyvale, CA 94089 \\
  \texttt{rabakshi@amazon.com} \\
  \And
  Kahkuen Fu\\
  Amazon Lab126 \\
  Sunnyvale, CA 94089 \\
  \texttt{kahkuen@amazon.com} \\
}

\begin{document}

\maketitle

\begin{abstract}
 Despite the proliferation of diverse hardware accelerators (e.g., NPU, TPU, DPU), deploying deep learning models on edge devices with fixed-point hardware is still challenging due to complex model quantization and conversion. Existing model quantization frameworks like Tensorflow QAT [1], TFLite PTQ [2], and Qualcomm AIMET [3] supports only a limited set of quantization schemes (e.g., only asymmetric per-tensor quantization in TF1.x QAT [4]). Accordingly, deep learning models cannot be easily quantized for diverse fixed-point hardwares, mainly due to slightly different quantization requirements. In this paper, we envision a new type of model quantization approach called MRQ (model re-quantization), which takes existing quantized models and quickly transforms the models to meet different quantization requirements (e.g., asymmetric → symmetric, non-power-of-2 scale → power-of-2 scale). Re-quantization is much simpler than quantizing from scratch because it avoids costly re-training and provides support for multiple quantization schemes simultaneously. To minimize re-quantization error, we developed a new set of re-quantization algorithms including weight correction and rounding error folding. We have demonstrated that MobileNetV2 QAT model [7] can be quickly re-quantized into two different quantization schemes (i.e., symmetric and symmetric+power-of-2 scale) with less than 0.64 units of accuracy loss. We believe our work is the first to leverage this concept of re-quantization for model quantization and models obtained from the re-quantization process have been successfully deployed on NNA in the Echo Show devices.
\end{abstract}

\section{Introduction}

With the popularity of deep neural networks across various use-cases (e.g., CV/AR, Audio/Speech, NLP), there is an increasing demand for methods that make deep networks run efficiently on resource-limited edge devices. One of most compelling methods is model quantization, where the float 32-bit weights and activations are quantized to 8-bit integer. Then, the quantized models can be run efficiently on fixed-point hardware with several benefits (i.e., ~2-3x faster inference, 4x less memory and bandwidth usage, less chip BOM cost and energy consumption).

There are many different types of quantization schemes for deep learning models (e.g., per-tensor vs. per-channel, symmetric vs. asymmetric, power-of-2 scale vs. non-power-of-2 scale) (refer to related work for details). Depending upon the specific hardware configuration and relative importance of computing efficiency, and accuracy different types of quantization schemes may be better suited for a particular task. For example, most of MCUs [5] only support symmetric quantization and some NPUs (e.g., Xillix [6]) requires power-of-2 scale quantization. Unfortunately, existing quantization frameworks and toolkits like Tensorflow QAT (Quantization-aware Training) [1], TFLite PTQ (Post-training Quantization) [2], and Qualcomm AIMET[3] do not support all types of quantizations. For example, only asymmetric per-tensor quantization is supported by TF1.x QAT[4] while most of frameworks do not support power-of-2 scale quantization at all. Due to such a mismatch between hardware and software quantization supports, it is challenging to quantize deep learning models for diverse fixed-point hardwares. Currently, quantizing deep learning models for diverse fixed-point hardware requires significant engineering effort to migrate to different training and quantization frameworks (or mix-and-match them), or even invent a new quantization algorithm for a specific hardware.

To address the challenge, we propose a new type of model quantization called MRQ (model re-quantization). It takes existing quantized models and quickly transforms the models to meet different quantization requirements (e.g., asymmetric → symmetric, non-power-of-2 scale → power-of-2 scale). Although hardware cannot run model unless exact quantization requirements are met, at a high level each quantization scheme is actually quite similar to each other. Thus, the re-quantization from existing quantized model is often simpler than quantizing from scratch. However, naive re-quantization (i.e., just adjusting quantization scale and zero point without any compensation) could introduce re-quantization error. To minimize this error, we developed a new set of algorithms including weight correction and rounding error folding. Thus, we could support multiple types of quantization schemes simultaneously without re-training. Through experiment, we demonstrated that MobileNetV2 QAT model [7] (asymmetric) can be quickly re-quantized into two different types of quantization schemes (i.e., symmetric and symmetric+power-of-2 scale) with less than 0.64\% accuracy loss (imagenet top1). Our contributions can be summarized as below.
\begin{itemize}
\item We envision MRQ which can re-quantize existing quantized model without costly re-training. 
\item Considering the characteristic of re-quantization, we develop weight correction and rounding error fold algorithms. 
\item We empirically validated that MRQ can re-quantize existing models into multiple quantization schemes with minimal accuracy loss.
\end{itemize}

\subsection{Related Work}

A good overview of deep learning model quantization is summarized in [8]. In general, we can further classify the quantization schemes (or requirements) into three categories. Each scheme has its own pros and cons.
\begin{figure}
  \centering
  \begin{minipage}{0.5\textwidth}
      \centering
      \includegraphics[width=0.99\textwidth]{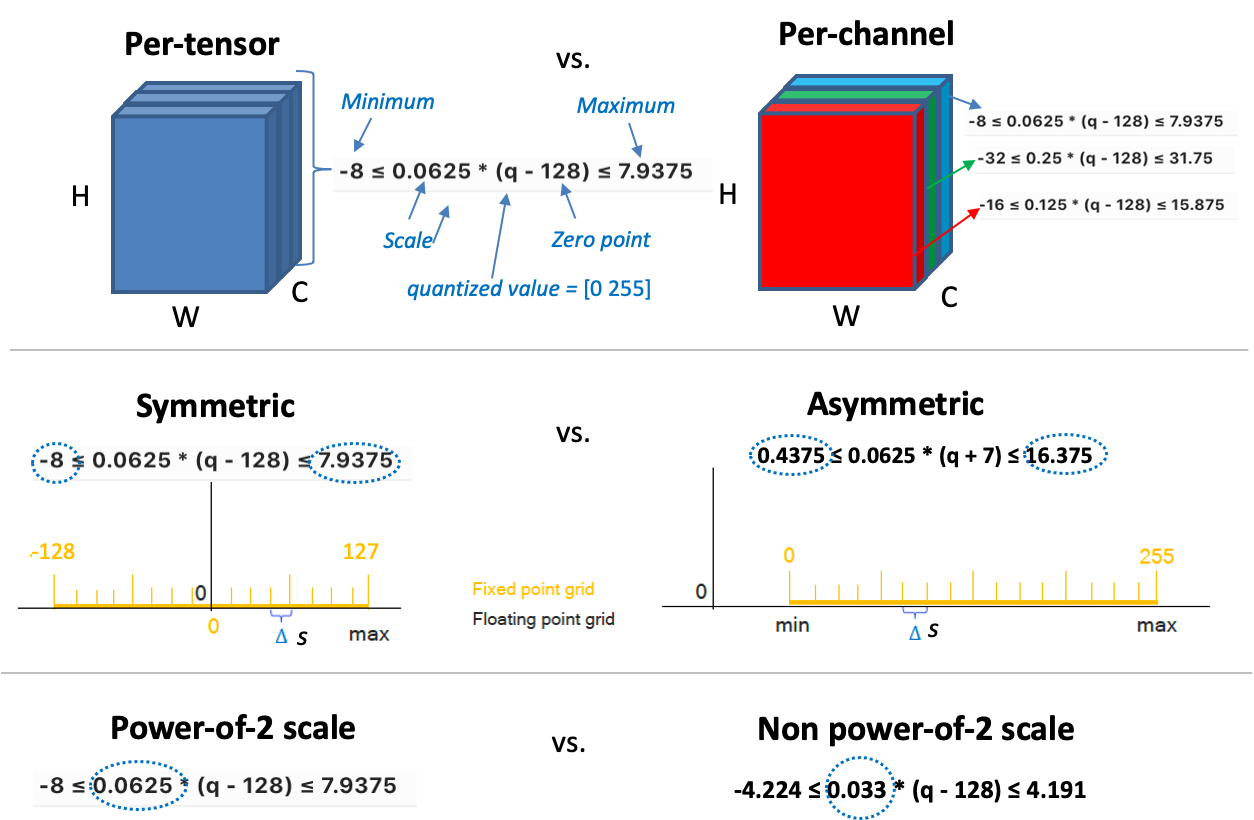} 
      \caption{Quantization Schemes}
  \end{minipage}\hfill

\end{figure}
\begin{table}

  \centering
  \begin{tabular}{lll}

    \cmidrule(r){1-3}
    Quantization schemes     & Computing efficiency     & Easy-to-quantize (accuracy)\\
    \midrule
   Per-tensor & Good  & No    \\
   Per-channel     & Bad & Yes      \\
    Symmetric     & Good       & No  \\
    Asymmetric & Bad & Yes     \\
    Power-of-2 scale     & Good & No      \\
    Non-power-of-2 scale     &  Bad    & Yes  \\
    \bottomrule
  \end{tabular}
  \\
    \caption{Quantization scheme comparison }
\label{tab:quant-table}
\end{table}

\begin{itemize}

\item Per-tensor vs. Per-channel: Per-tensor quantization has one quantization parameter set (i.e., scale and zero point) per weight and activation tensor. In contrast, per-channel quantization allows the use of different quantization parameter sets for each channel, facilitating quantization without accuracy loss. Typically, per-tensor quantization is better for computing efficiency in fixed-point hardware than per-channel quantization since a large set of tensor data can be processed together. However, per-tensor quantization can suffer from the divergence of dynamic range of each channel, hindering the ability to recover original accuracy back. Recently, more hardware and software frameworks support per-channel quantization additionally for convolution and depthwise-convolution due to its easy-to-quantization.

\item Symmetric vs. Asymmetric: It is classified by the symmetricity of dynamic range of tensor. Symmetric quantization is better for computing efficiency due to less auxiliary calculation. However, it can potentially cause more quantization noise due to limited expressiveness (e.g., [-6 6] for relu6 in symmetric quantization). TF 1.x QAT [4] supports only asymmetric quantization, but TF 2.x QAT [1] supports both symmetric and asymmetric quantizations. In our paper, we consider that the same symmetricity is applied into both weights and activations tensors.
\item Power-of-2 scale vs. Non-power-of-2 scale: Power-of-2 scale quantization is a special type of symmetric quantization, where scale parameter should be power-of-2. It is good for computing efficiency since re-scaling is simply done by bit-shifting instead of multiplication. Re-scaling is an operation after multiplying input and weight tensors. It adjusts the scale of multiply-accumulated results by considering the scales of input, weight and output tensors. However, power-of-2 scale also can introduce more quantization noise due to the rounding effects of dynamic range to power-of-2. Note that MRQ specially addresses this challenge through rounding error folding algorithm.
\end{itemize}

Model quantization from 32-bit floating-point to 8-bit fixed-point often results in the degradation of accuracy. To recover accuracy back to that of the original model, several model quantization methods like Quantization-Aware Training (QAT) or Post-Training Quantization (PTQ) have been proposed. PTQ methods [14, 15, 16, 17] identify the quantization parameters through weights and a small set of calibration dataset without full retraining. PTQ generally works well on 8-bit per-channel quantization, but does not work well for per-tensor quantization or needs more complicated algorithms [10, 11, 18]. In contrast, QAT methods [1, 4, 12, 13] directly trained the weights and quantization parameters by simulating quantization steps and thus generally outperforms PTQ in terms of accuracy with a cost of more training time and efforts. QAT is well supported by most of training frameworks like Tensorflow [1] and Pytorch [19]. Our MRQ is different from existing approaches in that the quantization starts from quantized models rather than original 32-bit models. For example, Tensorflow hub [9] already provides a lot of 8-bit quantized models, typically with QAT. MRQ could quickly transform the models into other models with different quantization schemes with minimal accuracy loss.

Table 2 summarizes quantization schemes supported by widely-used quantization frameworks. As shown in the table, most frameworks support only a limited set of quantization schemes. For example, power-of-2 scale quantization is only supported by Xilinx TQT [6] although TQT does not support generic models; only supports legacy TF v1.13 models. Since each framework supports different set of quantization schemes, if you want to quantize the model into different quantization schemes, you have to migrate the training pipeline to different training and quantization frameworks, or use multiple frameworks simultaneously to quantize the model from the scratch, requiring a lot of engineering efforts. Different from existing frameworks, MRQ could quickly support multiple quantization schemes in a unified way. For example, starting from most widely-used Tensorflow QAT model (protocol buffer format), we can modify the model with different quantization schemes and apply our algorithms to minimize re-quantization error. Then, the model can be converted to desired tflite model for edge devices by reusing Tensorflow-to-TFLite converter.
\begin{table}[]
\resizebox{\columnwidth}{!}{%
\begin{tabular}{lcccccccc}
\toprule
\thead{Type} & \thead{Framework} & \thead{Per-tensor\\ quantization} & \thead{Per-channel\\ quantization} & \thead{Symmetric} & \thead{Asymmetric} & \thead{Power-of-2 \\scale} & \thead{Non-power-of-2 \\scale} & \thead{Support\\generic models}  \\ 
\midrule
\multirow{4}{*}{QAT} & TF1.x QAT {[}4{]}                         &     \checkmark                     &                          &           &  \checkmark           &                  &       \checkmark                &      \checkmark                   \\
                     & TF2.x QAT {[}1{]}                         &     \checkmark                     &      \checkmark                     &   \checkmark         &    \checkmark         &                  &   \checkmark                    &    \checkmark                     \\
                     & IBM LSQ {[}13{]} / Qualcomm LSQ+ {[}12{]} &        \checkmark                    &                          &    \checkmark          &            &                  &      \checkmark                   &                        \\
                     & Xilinx TQT {[}6{]}                        &   \checkmark                         &                          &    \checkmark          &            &   \checkmark                  &                      &                        \\ 
\midrule
\multirow{3}{*}{PTQ} & TFLite PTQ {[}2{]}                        &                         &  \checkmark                          &\checkmark             &            &                  & \checkmark                       &   \checkmark                       \\
                     & Pytorch {[}19{]}                          &  \checkmark                         &    \checkmark                        &   \checkmark          &  \checkmark            &                  &    \checkmark                    &  \checkmark                        \\
                     & Qualcomm AIMET {[}3{]}                    &    \checkmark                       &       \checkmark                     &           & \checkmark             &                  &  \checkmark                      &   \checkmark                       \\ \cmidrule(l){1-9} 
\end{tabular}%
}
\\

\caption{Features available in open-source libraries}
\label{tab:my-table}
\end{table}

\section{Re-Quantization Algorithms}
\label{headings}

In re-quantization process the quantization parameters are updated to satisfy different quantization requirements than the original model had which are asymmetric → symmetric for all 3 tensors, and non-power-of-2 scale → power-of-2 scale. Satisfying these requirements simplifies the inference operation as described in equation 3. Implementation of such operation on hardware is very efficient as it only requires multiply and accumulate operation followed by bit shifts.
\begin{equation}
T = S(q - Z)
\end{equation}

\begin{equation}
S_o(q_o-Z_o) = \sum{S_w(q_w-Z_w)S_i(q_i-Z_i) }+ b
\end{equation}

\begin{equation}
S_o = 2^{-Q}(\sum{q_wq_i }+ b^*) \quad \textrm{where} \quad 2^{-Q} = \frac{S_wS_i}{S_o} \quad \textrm{and} \quad b^{*} = b( \frac{S_wS_i}{S_o} )^{-1}
\end{equation}
A naive way to meet these requirements is to ignore the zero points and make scale factor as the closest power of two. Such quantization puts the accuracy of the model close to random as shown in results section. For successful re-quantization there are four techniques which when used in conjunction gives the best overall accuracy. These techniques are bias correction [10, 15], weight clipping [10, 17], weight correction and round error folding. Bias correction and weight clipping are popular PTQ techniques but were found not sufficient in re-quantization, specifically to enable scaling as power of 2 constraints. Round error folding is a technique that we developed specifically to satisfy power of two constraints. 

\subsection{Bias correction}

Bias correction is a method to remove offset error due to quantization. A calibration dataset of size of 50-100s inputs is fed to original floating point model and the quantized model. The mean values of outputs of a layer are obtained. This value is the mean quantization error which is corrected for by adding to the layer’s bias. This is shown in eq 4. 
\begin{figure}
  \centering
  
   \includegraphics[width=5cm]{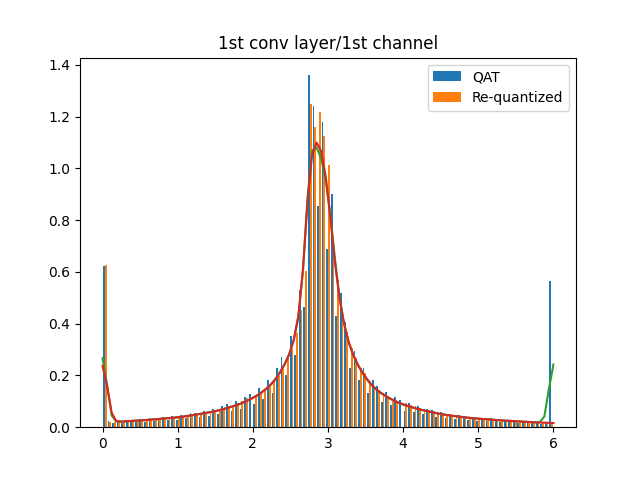}\hfil
    \caption{Distribution of Conv layer for a QAT model and requantized model.}
   \includegraphics[width=5.1cm]{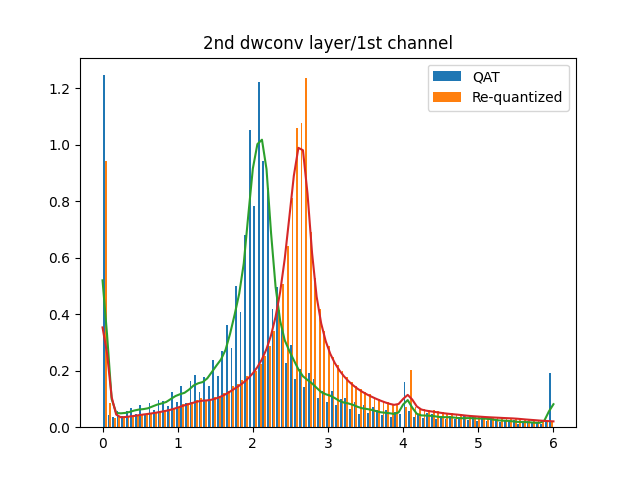}\hfil
    \caption{Distribution of Depthwise Conv layer for a QAT model and requantized model.}
 \end{figure}
\begin{equation}
b_c = b + \epsilon \quad \textrm{where} \quad \epsilon = E[T_o] - E[S_o(q_o - Z_o)]
\end{equation}

\subsubsection{Weights and Activation Clipping}

Clipping is method of setting a hard min and max limits for the floating point weights and activation of a layer before quantization. This technique is particularly useful when the tensor has few values which are outliers. A simple min max calculation to determine scales can get corrupted because of these outliers. Generally the choice of this clipping range  is set by visualizing and getting some statistics on the distribution of the weights. There are automated ways to identify the correct clipping range using MSE minimization, fitting gaussian or laplace distribution or minimizing KL divergence [17]. The choice of clipping determines the resolution that can be represented in integer representation. If dynamic ranges of weights and activation are below 6, each integer in 8bits (i.e., [0 255]) we could quite accurately represent float value (i.e., precision = 6/255 = 0.0235) Experiments show that choosing threshold beyond 6 is when the model’s accuracy starts to go down. Figure 4a shows the impact of clipping on quantized values.
\begin{equation}
t_c = min(max(t_,-c),c)  \quad \textrm{where} \quad t\epsilon T, t_c\epsilon T_c
\end{equation}

\begin{figure}
  \centering
  
   \includegraphics[width=5cm]{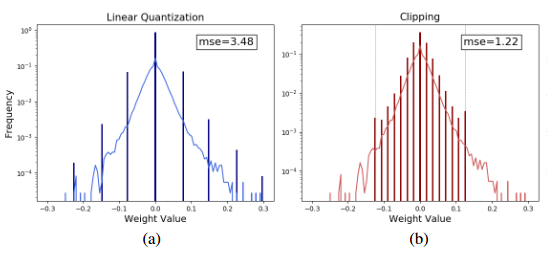}\hfil
    \caption{Figure 4a demonstrates quantized weight distribution with out clipping and 4b demonstrates with clipping.
}
   \end{figure}

\subsubsection{Weight Correction}
For QAT models, the scales and zero points for weights are calculated during forward pass in training time. So with re-quanization of a QAT model, if techniques like weight clipping are applied which would change scale, the quantization parameters originally obtained from QAT are not most optimal. Hence the floating point weights have to be re-quantized with the new quantization parameters. This is called weight correction. The floating point weights in QAT model before the fake quantization layer are called backward weights and obtained after the fake quantization layer are called forward weights. Since these forward weights are not valid because of change in quantization parameters, they are corrected by applying quantization as per the new parameters.
\begin{equation}
q_w = \lbrack[\Big] \frac{T_w}{S_w} \rbrack[\Big]  + Z_w \quad \textrm{where [v] is round, ceil or floor of v} \quad 
\end{equation}

\begin{equation}
q_w^{*} = \lbrack[\Big] \frac{T}{S_w^{*}}\rbrack[\Big]  +Z_w^{*} 
 \end{equation}
 
 \subsubsection{ Round Error Folding}
This is a technique which we developed to make the floating point multiplier a power of two. This makes multiplication efficient on HW using bit shift. Once the model is made symmetric, the inference operation as defined in eq 2 reduces to eq 8.
\begin{equation}
q_o = M (\sum{q_wq_i}+b^{*})  \quad \textrm{where} \quad M = \frac{S_wS_i}{S_o}  \quad \textrm{and} \quad  b^{*} = M^{-1}b
\end{equation}
To make the floating point multiplier M power of two, the quantized model is re-quantized. The M is decomposed into two values as given in eq 9. This P is the round error that gets folded into weights using weight correction method described in previous sub-section. The scale of weight tensor is modified for re-quantization as given in equation 10. 
\begin{equation}
M = P * 2^{-Q} \quad \textrm{where} \quad P \epsilon (0.5,1], Q \epsilon I
 \end{equation}
\begin{equation}
S_w^{*} = \frac{S_w}{P}
 \end{equation}
 Inference using re-quantized weight tensor qw* is defined in eq 11. The re-quantized weights makes the runtime floating point multiplier M* to be power of 2 as shown in eq 12. This can be implemented as right shifts if Q>0.
 \begin{equation}
q_o = M^{*}(\sum{q_w^{*}q_i + b^{**}})\quad \textrm{where} \quad M^{*} = \frac{S_w^{*}}{S_o} \quad \textrm{and} \quad b^{**} = M^{*-1}b
 \end{equation}
 \begin{equation}
M^{*} = \frac{S_w^{*}S_i}{S_o} = \frac{M}{P} = 2^{-Q}
\end{equation}
Important thing to note here is to get the value of P, scaling factors of the quantized model are used. Then this value is used to update the scale factor of weights which are re-quantized. This way this power of 2 requirement is met in the re-quantized model which was not in original quantized model.
\section{Experiments}
\label{others}
In this section, we evaluate the accuracy of our proposed re-quantization techniques. For this purpose, we test MobilenetV2 model. MobilenetV2 is a representative backbone network for diverse computer vision recognition tasks (e.g., semantic segmentation, classification, and object detection). It is widely used on edge devices due to light-weight computation along with a relatively good accuracy. The model is pre-trained on the ImageNet dataset and hosted by tensorflow lite [20]. The models were evaluated on the entire ImageNet Validation dataset (i.e., 50,000 images).
\begin{figure}
  \centering
  
   \includegraphics[width=5cm]{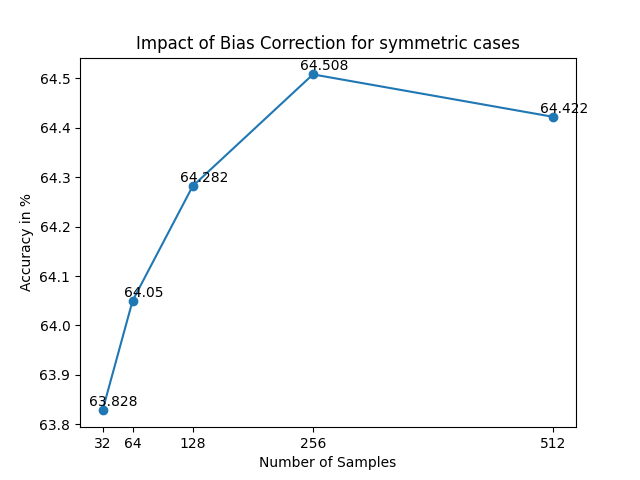}\hfil
   \caption{Impact of bias correction for symmetric cases}   
   \includegraphics[width=5.1cm]{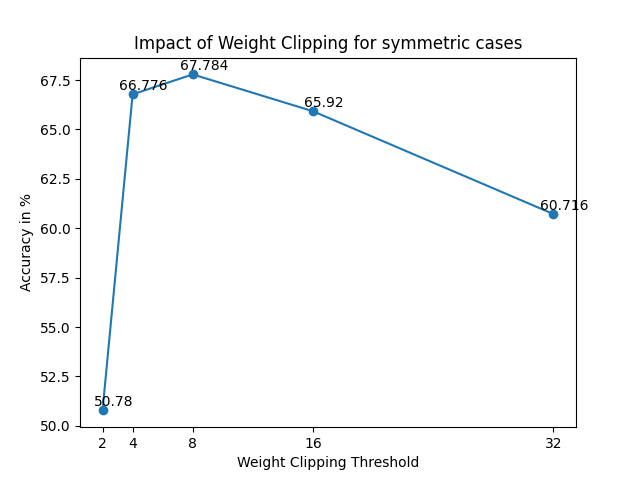}\hfil
    \caption{Impact of weight clipping for symmetric cases}   
 \end{figure}
\begin{table}

  \centering
  \begin{tabular}{lll}

    \cmidrule(r){1-3}
    Requantization schemes     & Symmetric     & Symmetric and Power-of-2 scale\\
    \midrule
    Original (QAT) & 70.64  & 70.64     \\
    Naive     & 5.44 & 11.00      \\
    BC     & 64.28       & 63.95  \\
    BC + WCL & 67.78 & 65.97     \\
    BC + WCL + WCR     & 70.31 & 64.90      \\
    BC + WCL + WCR + REF     &  ---     & 70.00  \\
    \bottomrule
  \end{tabular}
  \\
    \caption{Experimental results}
\label{tab:exp-table}
\end{table}

Our experiments will demonstrate that MobileNetV2 can be quickly re-quantized into two different quantization schemes: 1) symmetric quantization (e.g., mostly used by MCUs [5]), 2) symmetric quantization with power-of-2 scale constraints (e.g., Xilinx NPUs [6]). Our re-quantization technique involves in manipulating the values obtained by Quantization Aware training of minimum and maximum in the operation ‘FakeQuantWithMinMaxVars’. For re-quantization, we adjust them as follows:
\begin{itemize}
\item Symmetric: maximum'= max(abs(minimum), abs(maximum)), minimum’ = -maximum’
\item  Symmetric + power-of-2 scale: maximum’ = power(2, round(log2(max(abs(minimum), abs(maximum))))), minimum’ = -maximum’
\end{itemize}
The following experiments were conducted:
\begin{itemize}
\item  Bias Correction (BC)
\item  Bias Correction and Weight Clipping (BC+WCL)
\item  Bias Correction and Weight Clipping with Weight Correction (BC + WCL + WCR)
\item Bias Correction and Weight Clipping with Weight Correction and Round Error Folding (BC + WCL + WCR+ REF)
\end{itemize}
Overall, we observed that quantizing for both symmetric and power-of-2 constraints is harder than just symmetric. The existing techniques of PTQ (Bias correction and Weight Clipping) was insufficient in regaining the accuracy. Our proposed techniques (Weight Correction for symmetric quantization and Round error folding for symmetric and power-of-2) were effective to fully recover the accuracy for the presented criterion. Round Error Folding proved to be effective for symmetric and power-of-2 cases with almost 6\% improvement in accuracy. It was also observed that Weight Correction was effective for symmetric-only cases but ineffective for symmetric and power-of-2.

\section{Conclusion and Future work}
In this paper, we envisioned new type of model quantization called MRQ (model re-quantization). Different from existing quantization approaches like QAT and PTQ, MRQ starts quantization from existing quantized models rather than original 32-bit floating point models, which is often simpler than quantizing from the scratch. MRQ does not require any costly re-training and easy-to-support multiple types of quantization schemes simultaneously. To minimize re-quantization error, we developed a  new set of re-quantization algorithms by fully considering the characteristic of re-quantization. Also, we demonstrated that MobileNetV2 QAT model [7] can be quickly re-quantized into two different quantization schemes (i.e., symmetric and symmetric+power-of-2 scale) with less than 0.64\% accuracy loss. As a future direction, we plan to apply our re-quantization approach into more complex deep learning models like object detection and semantic segmentation.

\section*{References}

\small

[1] Google Tensorflow QAT (Quantization-Aware Training), https://www.tensorflow.org/model\_optimization/guide/quantization/training

[2] Google TFLite PTQ (Post-training Quantization), https://www.tensorflow.org/lite/performance/post\_training\_quantization

[3] Qualcomm AIMET (AI Model Efficiency Toolkit), https://github.com/quic/aimet

[4] Object Detection API with TensorFlow 1, https://github.com/tensorflow/models/blob/master/research/object\_detection/g3doc/tf1.md

[5] ARM CMSIS-NN, https://github.com/ARM-software/CMSIS\_5/tree/develop/CMSIS/NN

[6] Jain, Sambhav, et al. {\it {"Trained quantization thresholds for accurate and efficient fixed-point inference of deep neural networks." Proceedings of Machine Learning and Systems 2 (2020): 112-128.}} s (https://github.com/Xilinx/graffitist)

[7]  TFLite hosted models, https://www.tensorflow.org/lite/guide/hosted\_models

[8]  Nagel, Markus, et al.  {\it {"A white paper on neural network quantization."arXiv preprint arXiv:2106.08295}}(2021)

[9] Tensorflow hub, https://www.tensorflow.org/hub

[10] Nagel, Markus, et al.{\it {"Data-free quantization through weight equalization and bias correction." Proceedings of the IEEE/CVF International Conference on Computer Vision. 2019.}}

[11] Nagel, Markus, et al.{\it { "Up or down? adaptive rounding for post-training quantization." International Conference on Machine Learning. PMLR, 2020.}}

[12] Bhalgat, Yash, et al. {\it {"Lsq+: Improving low-bit quantization through learnable offsets and better initialization." Proceedings of the IEEE/CVF Conference on Computer Vision and Pattern Recognition Workshops. 2020.}}

[13] Steven K Esser, Jeffrey L McKinstry, Deepika Bablani, Rathinakumar Appuswamy, and Dharmendra S Modha. {\it {"Learned step size quantization". arXiv preprint arXiv:1902.08153, 2019 }}

[14] Xu, Suwa, Jinwon Lee, and Jim Steele. {\it{"PSVD: Post-Training Compression of LSTM-Based RNN-T Models." 2021 IEEE Automatic Speech Recognition and Understanding Workshop (ASRU). IEEE, 2021.}}

[15] Finkelstein, Alexander, Uri Almog, and Mark Grobman. {\it {"Fighting quantization bias with bias." arXiv preprint arXiv:1906.03193 (2019).}}

[16] Ron Banner, Yury Nahshan, and Daniel Soudry. {\it {"Post training 4-bit quantization of convolutional networks for rapiddeployment". In Advances in Neural Information Processing Systems, pages 7948\textendash{}7956, 2019.}}

[17] Ritchie Zhao, Yuwei Hu, Jordan Dotzel, Chris De Sa, and Zhiru Zhang. {\it {"Improving neural network quantization without retraining using outlier channel splitting". In International Conference on Machine Learning, pages 7543\textendash{}7552, 2019.}}

[18] Yaohui Cai, Zhewei Yao, Zhen Dong, Amir Gholami, Michael W. Mahoney, and Kurt Keutzer. {\it {"Zeroq: A novel zero shot quantization framework." CoRR, abs/2001.00281, 2020}}

[19] Pytorch quantization, https://pytorch.org/docs/stable/quantization.html

[20] MobilenetV2 hosted on Tensorflow Hub: https://storage.googleapis.com/download.tensorflow.org/models/tflite\_11\_05\_08/mobilenet\_v2\_1.0\_224\_quant.tgz

[21] Jacob, Benoit, et al.{\it{ "Quantization and training of neural networks for efficient integer-arithmetic-only inference." Proceedings of the IEEE conference on computer vision and pattern recognition. 2018}} 
\end{document}